\documentclass[11pt]{article}
\usepackage{arxiv}
\usepackage{listings}
\usepackage{todonotes}
\usepackage[all, cmtip]{xy}
\usepackage{subcaption}
\usepackage{amsfonts, amsmath}
\usepackage{diagxy}
\usepackage{hyperref}
\usepackage{booktabs}
\usepackage{adjustbox}
\usepackage{multirow}
\usepackage[numbers]{natbib}
\usepackage{tabularx}
\newcommand{\fst}[1]{\textbf{#1}}

\title{From axioms over graphs to vectors, and back again: evaluating
  the properties of graph-based ontology embeddings}

\author{
  Fernando Zhapa-Camacho, Robert Hoehndorf \\
  Computational Bioscience Research Center,\\ Computer, Electrical \&
  Mathematical Sciences and Engineering Division \\
  King Abdullah
  University of Science and Technology, 4700 KAUST, 23955 \\
  Thuwal, Saudi Arabia \\
  \texttt{\{fernando.zhapacamacho,robert.hoehndorf\}@kaust.edu.sa}
}

\date{}

\begin{document}
\maketitle

\begin{abstract}
  Several approaches have been developed that generate embeddings for
  Description Logic ontologies and use these embeddings in machine
  learning. One approach of generating ontologies embeddings is by
  first embedding the ontologies into a graph structure, i.e.,
  introducing a set of nodes and edges for named entities and logical
  axioms, and then applying a graph embedding to embed the graph in
  $\mathbb{R}^n$.  Methods that embed ontologies in graphs (graph
  projections) have different formal properties related to the type of
  axioms they can utilize, whether the projections are invertible or
  not, and whether they can be applied to asserted axioms or their
  deductive closure.  We analyze, qualitatively and quantitatively,
  several graph projection methods that have been used to embed
  ontologies, and we demonstrate the effect of the properties of graph
  projections on the performance of predicting axioms from ontology
  embeddings. We find that there are substantial differences between
  different projection methods, and both the projection of axioms into
  nodes and edges as well ontological choices in representing
  knowledge will impact the success of using ontology embeddings to
  predict axioms.
\end{abstract}


\section{Introduction}
\label{sec:intro}

Ontologies are widely used to integrate and standardize data across
databases.  In recent years, ontologies and their associated knowledge
in databases have been used in machine learning to constrain the
solution space by the ontology structure, with several applications in
the biomedical domain~\cite{semantic_similarity}.
One form of using ontologies in machine learning tasks is based on
graphs \cite{onto2graph, dl2vec, owl2vecstar}. We use the term
\textit{graph projection} to refer to the transformation of an
ontology into a graph. With recent developments in machine learning
over graphs \cite{kgsurvey}, several approaches to project ontologies
into graphs have emerged. These approaches are able to capture the
ontology structure at some level and have been evaluated on tasks such
as similarity computation, link (axiom) prediction \cite{owl2vecstar},
or ontology alignment \cite{owl2vec_align}.

Graphs as a form of representation of knowledge have been studied for
several decades \cite{Dau_2001, logicGA}. Existential Graphs (EGs)
were proposed by C. S. Peirce as a means to depict logical expressions
through diagrams. EGs enabled the representation of first order logic formulas
\cite{pierce_tutorial}. Semantic Networks, like EGs, are graphs which
contain representations of concepts as nodes and relations between
concepts as edges \cite{what_is_a_link, what_is_a_sn}. Conceptual
graphs arose from both EGs and Semantic Networks, by leveraging the
logical foundation of EGs with the properties of Semantic Networks
\cite{from_eg_to_cg}. The diagrammatic representation of logical
expressions has not only been developed to make such expressions more
readable and understandable, but also to enable certain operations,
for example those that correspond to forms of computing entailments
and reasoning. In \citet{dau_2005}, the formalization of the diagrams
of Existential Graphs and their use in logic calculus and reasoning is
explored. Further work \cite{diagrammatic_alc, conceptual_ontology}
investigates the reasoning capabilities that existential and
conceptual graphs can have for Description Logics.

There has been a renewed interest in graph-based representations of
ontologies with the emergence of graph-based machine learning
methods. Graph embedding methods and knowledge graph embeddings
\cite{kgsurvey} have been developed to embed (knowledge) graphs in the
$\mathbb{R}^n$ where gradient-based methods can be used to solve
optimization problems that allow these embeddings to be used, for
example, for the prediction of edges or for determining similarity
between nodes. A graph projection embeds an ontology in a graph
structure; this graph structure can then be used to generate
embeddings of ontology entities in $\mathbb{R}^n$. Because graphs can
be intermediate steps in generating ontology embeddings (in
$\mathbb{R}^n$), it becomes important to investigate properties of
graph projections.  A graph projection is \textit{total} if it uses of
all the axioms in the ontology to generate a graph, and
\textit{partial} otherwise. Totality can be defined with respect to
asserted axioms in an ontology, or with respect to the deductive
closure.  Relating a (predicted) edge in a graph to an axiom in an
ontology requires that the graph projection is \emph{injective} (i.e.,
that different axioms induce different subgraphs). This is not true
for every graph projection method. Furthermore, graph projections can
project axioms into single edges or into subgraphs (potentially with
multiple edges); some graph embedding methods are designed to predict
single edges and not subgraphs, and graph projections that generate
subgraphs may therefore not be suitable to be used jointly with those
graph embeddings. Analyzing these properties is crucial to
understanding how the information provided by ontologies is utilized
by each method and what their limitations are; understanding the
limitations enables the development of new methods that can address
them. While graph-based embeddings are not the only way to embed
ontologies, graphs are widely used due to the large availability of
graph-based machine learning methods.  Here, we analyze graph
projections and their properties in the context of embedding
ontologies in $\mathbb{R}^n$. Our contributions are the following:
%\begin{itemize}
(a) We provide a qualitative analysis of graph projection methods with
respect to totality, injectivity, and use of deductive closure; (b) We
quantitatively analyze the effect of properties of graph projections
in experiments that predict axioms in the deductive closure of
ontologies.

\section{Ontology embeddings and graph embeddings}
 
\begin{figure}[t]
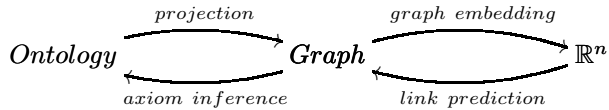

  \centering
  \[
    \bfig
    \morphism(0,0)|a|/{@{>}@/^1em/}/<1000,0>[Ontology`Graph;projection]
    \morphism(0,0)|b|/{@{<-}@/_1em/}/<1000,0>[Ontology`Graph;axiom\ inference]
    \morphism(1000,0)|a|/{@{>}@/^1em/}/<1000,0>[Graph`\mathbb{R}^n;graph\ embedding]
    \morphism(1000,0)|b|/{@{<-}@/_1em/}/<1000,0>[Graph`\mathbb{R}^n;link\ prediction]
    \efig
  \]
  \caption{Ontology embedding (left to right) and inference (right to left) processes.}\label{fig:embed_process}
\end{figure}

An embedding is a structure-preserving mapping between two
mathematical structures.
A graph-based embedding is a
two-step process where an ontology is first embedded into a graph, and
then a graph-embedding is used to embed the graph in the
$\mathbb{R}^n$.  We call the first embedding an ``graph
projection'' or ``projection'' and the second embedding a ``graph
embedding''.
Within $\mathbb{R}^n$, inferences may then be done approximately
and translated back to the ontology. The inference computation is done
by computing plausibility of edges (links) in the graph that were
generated by a query axiom (Figure~\ref{fig:embed_process}).

While the machine learning and the Semantic Web communities have spent
substantial effort on designing methods that achieve the second part
(graph embeddings), the first part (graph projection) remains
rather unexplored. However, the property of the graph projection
itself has consequences for the types of operations that can be
performed in the embedding space ($\mathbb{R}^n$).
The main question that we investigate is how the mathematical
properties of graph projections affect the inference task. We
analyze totality and injectivity, as well as, the use of semantic
information in the process of graph generation.

\section{Graph projection methods}
\label{sec:qual}
\subsection{Preliminaries}

An ontology $\mathcal{O} = (\Sigma, Ax)$ is a tuple consisting of a
signature $\Sigma$ and a set of (Description Logic) axioms $Ax$.  The
signature $\Sigma = (\mathbf{C}, \mathbf{R}, \mathbf{I})$ consists of
a set of class names $\mathbf{C}$, a set of role names $\mathbf{R}$, a
set of individual names $\mathbf{I}$. The set $Ax$ is a set of
formulas in a language $\mathcal{L}(\Sigma)$.  We consider here only
ontologies where the set of axioms $Ax$ are formulated in a
Description Logic \cite{DLhandbook} language (see Appendix
\ref{app:ontologies}).  The deductive closure $\mathcal{O}^\vdash$ of
$\mathcal{O}$ is defined as
$\mathcal{O}^\vdash = \{\phi | Ax \vdash \phi \}$.

Relational graphs are intermediate structures during the ontology
embedding process.  A relational graph $G$ is a triple $(V,E,L)$,
where a $V$ is a set of vertices, $L$ is a set of edge labels, and
$E \subseteq V \times L \times V$ is a set of edges between vertices
and with a label from $L$.

The prediction of edges and the prediction of axioms can be formulated
as ranking problems where edges and axioms are scored using a scoring
function, with the intended meaning that a higher-scoring edge or
axiom is preferred over a lower-scoring edge or axiom.

A graph projection for $\mathcal{O} = (\Sigma, Ax)$ is a function
$p$ that maps an ontology into a relational graph $G = (V,E,L)$ such
that $\mathbf{C} \cup \mathbf{R} \cup \mathbf{I} \subseteq V \cup L$ (i.e., each class, individual, and
role name is represented as a node or edge label in $G$) and
for each $a \in Ax$, $p(a) \subseteq E$ (i.e., $p$ maps an axiom onto
a subgraph of $G$).  The function $p$ may be
total or partial with respect to $\mathcal{O}$ depending on whether it
is defined for all axioms in the set $Ax$ or only for some axioms. $p$
may also be total or partial with respect to the deductive closure of
$\mathcal{O}$ based on whether it is defined for all axioms in
$\mathcal{O}^\vdash$. We call $p$ simple if the cardinality of $p(a)$
is $1$ for all axioms $a$, i.e., if the projection function maps each
axiom onto exactly one edge.  $p$ takes an axiom as argument and
generates a graph. If $p$ is injective, it will generate different
subgraphs for different axioms and the projection function is
therefore invertible, i.e., from a (predicted) subgraph it becomes
possible to generate a corresponding axiom.

Here, we analyze different projection methods developed for ontologies
and their use in machine learning: (i)
taxonomic projection, OWL2Vec* and DL2Vec \cite{owl2vecstar, dl2vec},
Onto2Graph \cite{onto2graph}, and the graphs constructed from the RDF
rendering of OWL.
Examples of graphs generated by each projection method can be found in
Appendix \ref{app:graphs}.

\subsection{Taxonomy projection}

A taxonomy projection generates a graph from subclass axioms between
named classes. From ontology $\mathcal{O}$, the axioms used are those
of the form $C \sqsubseteq D$ where $C, D$ are class names; the projection
function is simple and generates a single edge, from $C$ to $D$.  The
projection is partial if $\mathcal{O}$ contains axioms besides
$C \sqsubseteq D$, and total otherwise; if it is total for $O$, the
projection is also total for $\mathcal{O}^\vdash$.  Furthermore, this
projection is both simple and injective, which means we can infer a
single axiom from a predicted edge.

\subsection{OWL2Vec*}
\label{sec:owl2vec}

OWL2Vec* \cite{owl2vecstar} (and variants such as DL2Vec \cite{dl2vec}) targets the
Description Logic $\mathcal{SROIQ}$ \cite{sroiq} underlying OWL 2 DL
\cite{owl2}.  The projection rules for the graph component in OWL2Vec*
are shown in Appendix \ref{app:owl2vec}.
In addition to the taxonomic structure, the OWL2Vec* projection
includes projections for axioms involving complex class descriptions,
including quantifiers and roles. The role names used with quantifiers
are used as labels in the relational graph.
For example, an axiom of the form
$Parent \sqsubseteq \exists hasChild. Person$ is transformed
into the edge $(Parent, hasChild, Person)$, which relates two nodes
using a labeled edge corresponding to the role $hasChild$. % In
The OWL2Vec* projection does not differentiate between quantifiers,
i.e., 
$p_{owl2vec*}(A \sqsubseteq \exists R.B) = p_{owl2vec*} (A \sqsubseteq
\forall R.B) = \{(A,R,B)\}$. Similarly, union ($\sqcap$) and
intersection ($\sqcup$) operators are not distinguished, i.e., 
$p_{owl2vec*}(A \sqsubseteq \exists R .( B \sqcap C)) = p_{owl2vec*} (A \sqsubseteq
\exists R. (B \sqcup C)) = \{(A,R,B), (A,R,C)\}$.
The OWL2Vec* projection is a partial function in both the set of
axioms $Ax$ and the deductive closure $\mathcal{O}^{\vdash}$ because
concept descriptions including operators such
as negation ($\neg$) are not defined for $p_{owl2vec*}$.
The OWL2Vec* projection is not injective because different axioms will
produce the same edge, such as when quantifiers are not
distinguished. 
When inferring an axiom from a graph, 
several axioms would obtain exactly the same score because they
generate the same edge or set of edges (and therefore the inverse of
$p_{owl2vec*}$ produces a set of axioms instead of a single
axiom). For example, if we query axioms $A \sqsubseteq \exists R. B $
or $A \sqsubseteq \forall R. B$, both will receive the score given to
the edge $(A, R, B)$. Moreover, $p_{owl2vec*}$ is not simple; for
example, the cardinality of  $p_{owl2vec*}(A \sqsubseteq \exists R. (B
\sqcap C))$ is greater than $1$ because the projection maps to edges
$\{(A,R,B),(A,R,C)\}$ (Figure~\ref{fig:owl2vec}).

\subsection{Syntax trees and RDF graphs}

We can use the syntactic representation of $Ax$ directly to generate
graphs using, for example, syntax trees.  One option is to use the
graph-based rendering of the OWL syntax in RDF \cite{rdf_owl2}, which is a representation of the syntax tree
underlying the Description Logic axioms in OWL.  The main advantage of
using a syntax tree as a graph representation is that the totality of the projection function are
guaranteed, both for axioms in $\mathcal{O}$ and the deductive closure
$\mathcal{O}^\vdash$. However, nodes in the relational graph
generated from the projection no longer correspond to named entities
in the signature of $\mathcal{O}$ (due to the introduction of internal
nodes in the syntax tree, or blank nodes in RDF).
For example, to represent the axiom $C \sqcap D  \sqsubseteq
\bot$, four blank nodes are created (Appendix Figure~\ref{fig:axiom1}).
Similarly, to represent the axioms $A \sqsubseteq \exists R.(B\sqcap C)$, five blank nodes are introduced (Figure~\ref{fig:rdf}). A major difference to methods such as OWL2Vec* is that
single axioms do not correspond to a single edge but rather to a
subgraph, i.e., the projection is, in general, not simple.
This raises an issue during axiom inference when axioms need to be
generated and scored, because the score will be computed from
subgraphs instead of single edges. 

\subsection{Relational axiom patterns}
\label{sec:onto2graph}

Onto2Graph \cite{onto2graph} is a method that implements graph
projection based on (relational) ontology design patterns
\cite{relations_as_patterns}.  In the past, ontologies in the
biomedical domain were often represented as directed
acyclic graphs \cite{obofoundry} and not using a formal language based
on a model-theoretic semantics. It took several years before the
graph representation of the ontologies was put on a formal
semantic foundation \cite{obo_syntax_semantics, owl1_1,
  relations_as_patterns}. Two approaches provided this
foundation, one based on a correspondence between edges and axioms of
a certain type as in the OWL2Vec* projection
\cite{obo_syntax_semantics, owl1_1}, and others based on relational
ontology design patterns \cite{relations_as_patterns}. The OBO
Relation Ontology \cite{Smith2005} contains a large number of such
patterns used in biomedical ontologies.

\begin{figure}[ht]
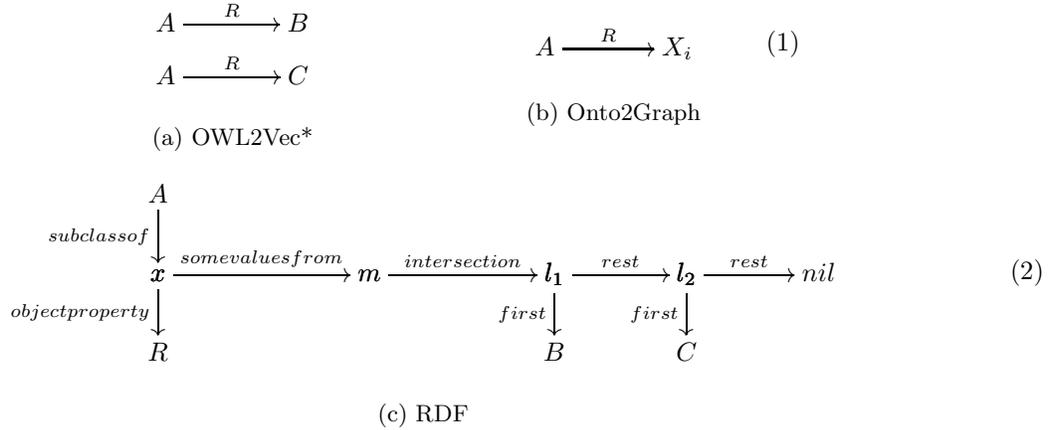

\centering
\begin{subfigure}{.3\linewidth}
  \centering
    \begin{equation*}
    \bfig
    \morphism[A`B;R]
    \morphism(0,-200)[A`C;R]
    \efig
  \end{equation*}
  \caption{OWL2Vec*}
  \label{fig:owl2vec}
\end{subfigure}
\begin{subfigure}{.3\linewidth}
  \centering
  \begin{equation}
    \bfig
    \morphism[A`X_i;R]
    \efig
  \end{equation}
  \caption{Onto2Graph}
  \label{fig:onto2graph}
\end{subfigure}
\bigskip
\begin{subfigure}{\linewidth}
  \centering
   \begin{equation}
    \bfig
    \morphism(-1000,0)<800,0>[x`m;somevaluesfrom]
    \morphism(-200,0)<700,0>[m`l_1;intersection]
    \morphism(500,0)<500,0>[l_1`l_2;rest]
    \morphism(1000,0)<500,0>[l_2`nil;rest]
    \morphism(-1000,300)<0,-300>[A`x;subclassof]
    \morphism(-1000,0)<0,-300>[x`R;objectproperty]
    \morphism(500,0)<0,-300>[l_1`B;first]
    \morphism(1000,0)<0,-300>[l_2`C;first]
    \efig
  \end{equation}
  \caption{RDF}
  \label{fig:rdf}
\end{subfigure} 
\caption{Graphs generated by each projection method on the axiom $A
  \sqsubseteq \exists R. (B\sqcap C)$. For the case of Onto2Graph, we
  can query every class $X_i\equiv B \sqcap C$ and generate the
  corresponding edge.}\label{fig:axiom2}
\end{figure}

A relational pattern is defined using variables that stand for symbols
and are used to define edges in a graph. An example of a relational
pattern is $?X \sqsubseteq \exists ?R.?Y$ from which an edge
$(?X, ?R, ?Y)$ can be created in a graph. More commonly, patterns that
use specific roles are used, such as $?X \sqsubseteq \exists
\mbox{part-of}.?Y$ to create an edge labeled ``part-of'' from $?X$ to
$?Y$. Relational patterns are flexible and can be defined for
arbitrary axioms. For example, a set of ``disjointness'' edges can be
created from an axiom pattern such as $?X \sqcap ?Y \sqsubseteq
\bot$. In Figure~\ref{fig:onto2graph}, the pattern $A\sqsubseteq
\exists R. ?X$
% \todo{Wrong syntax, missing quantifier!}
is selected, where $?X \equiv B \sqcap C$ is the query to apply to the ontology.
Given a relational pattern, an ontology can be queried for pairs or triples that satisfy these
patterns in quadratic ($O(| \mathbf{C}|^2)$)
% \todo{Not C --- cardinality of C!!!}
or cubic ($O(|\mathbf{C}|^2 |\mathbf{R}|)$)
% \todo{Cardinality}
time, respectively, by substituting every class name
and role name in the variables of the relational patterns. This
querying can be applied to either the set of axioms $Ax$ or their
deductive closure. The Onto2Graph \cite{onto2graph}
method implements an algorithm to generate graph edges $(?X, R,
?Y)$
%\todo{Should be ?Y (question mark missing)}
more efficiently for some axiom patterns.

$p_{onto2graph}$ is a partial function unless a pattern for every type
of axiom is defined (and the patterns are only generated from asserted
axiom and not their deductive closure). $p_{onto2graph}$ may be
injective if the patterns are defined so that different axioms map to
different subgraphs. However, the OWL2Vec* projection function can be
seen as a special case of relational patterns (where no domain
knowledge is used to specify patterns), and since the OWL2Vec*
projection is not injective, relational patterns may also not be
injective. In general, $p_{onto2graph}$ is not simple as multiple
edges can be generated from a single axiom; however, in practice, the
Onto2Graph projection is usually simple (i.e., the library of
relational patterns defined in the OBO Relation Ontology
\cite{Smith2005} implements only simple projections).

\section{Machine learning with graph projections}

\subsection{Queries and axiom scoring}

The main reason we investigate graph projections is due to the availability of
machine learning methods for graphs. In particular, (knowledge) graph
embeddings can be used for tasks such as determining similarity
between nodes \cite{dl2vec}, to predict edges that may be added to a
knowledge graph \cite{kg_completion}, or (approximately) answer complex queries
corresponding to subgraphs of the knowledge graph \cite{betaE, gammaE, tar}. 
Here, we investigate the impact of the properties of graph
projections 
on machine learning with ontologies. Specifically, for graph edges
$(h,r,t)$ that can be added to the graph,
knowledge graph embeddings can be used to define scores ($score(h,r,t)$), and we can use
$score(h,r,t)$ to score axioms that may be added to ontology
$\mathcal{O}$ (``axiom inference'').

The main operation that allows us to score and infer axioms is the
inverse of the projection function $p$. If $p$ is simple and
injective, its inverse $p^{-1}$ will yield exactly one axiom $a$ for
an edge $(h,r,t)$, and we can define $score(a) := score(h,r,t)$ to
score axioms; we also refer to scoring an axiom as a ``query''.
For example, to query the axiom $C \sqsubseteq \forall R.D$, we first
project the axiom onto a graph edge (for
example, the edge $(C, R, D)$ using the OWL2Vec* projection); then, we
determine the score of the edge $(C, R, D)$ using a knowledge graph
embedding method; and, finally, we apply the inverse of the projection
function to determine $score(C \sqsubseteq \forall R.D)$. If the
projection is not injective (such as the OWL2Vec* projection),
multiple axioms are generated with the same score; if axioms are added
to ontology $\mathcal{O}$ by ranking their scores, this needs to be
considered. For example, the inverse of the OWL2Vec* projection for
the edge $(C, R, D)$ will produce a set of axioms containing at
least $C \sqsubseteq \forall R.D$ and $C \sqsubseteq \exists R.D$,
with the same score.

Non-simple projections produce multiple edges for single axioms, and
computing the inverse requires determining a score for a
subgraph and transferring it to the axiom. While there
are methods to directly score subgraphs using knowledge graph
embeddings, we will use the arithmetic mean of the scores of each
edge in the subgraph as the score of the subgraph.

\subsection{Experimental setup}
\label{sec:exp}

To test the performance of different projection methods, we test their
ability to predict axioms in the deductive closure of an ontology. We
evaluate axiom prediction in two different ways: (i) we generate
embeddings using the original ontology ($\mathcal{O}$), and (ii) we
generate embeddings from a reduced version of the ontology
($\mathcal{O}_{reduced}$) by removing some axioms. The test set
consists of axioms that are in $\mathcal{O}^{\vdash}$ but not in
$\mathcal{O}^{\vdash}_{reduced}$.
The first case corresponds to computing entailments analogously to an
automated reasoner, and the second case corresponds to ``prediction''
of statements that may hold true although they are not entailed; the
second case may also be considered a form of approximate entailment
\cite{falcon}.

We used two large biomedical ontologies for evaluation, the Gene
Ontology \cite{go} (GO) and the Food Ontology (FoodOn)
\cite{foodon}. In GO, we tested prediction of subclass
($sub$) axioms of the form $C \sqsubseteq D$ and axioms involving
existential restrictions ($ex$) of the form
$C \sqsubseteq \exists R. D$. We generated reduced ontologies
GO$_{sub}$ and GO$_{ex}$ by randomly removing $10\%$ of the $sub$ and
$ex$ axioms, respectively.
In contrast to GO, FoodOn axioms use both quantifiers ($\exists,
\forall$). We used FoodOn to investigate the effect of injective
projections by testing the methods on the axiom patterns $C
\sqsubseteq \exists R.D$  and $C \sqsubseteq \forall
R.D$. Furthermore, to obtain a substantial difference between the
FoodOn$^{\vdash}$ and  FoodOn$^{\vdash}_{sub\_ex}$, we randomly
removed $30\%$ of the $sub$ and $ex$ axioms from FoodOn.  To generate
the deductive closure for all the ontologies and their reduced
versions, we used the OWLAPI and the ELK reasoner
\cite{elk}. ELK is a reasoner for OWL 2 EL ontologies. Other reasoners can be used, especially for FoodOn, where axioms can belong
to more complex description logics than $\mathcal{EL}$
\cite{DLhandbook}. However, due to the complexity of reasoning in
expressive description logics, we use ELK for our
experiments. Projection methods can handle axioms involving
existential and universal quantifiers,
% \todo{quantifiers?}
but axioms in the deductive closure will not involve universal restrictions due to the use of ELK.

To embed the graphs generated from projection methods,
we first use the knowledge graph embedding method TransE \cite{transe} (see
Appendix~\ref{app:kge} for details). While there are many methods to
embed knowledge graphs \cite{kgsurvey}, we use TransE because it is
a simple approach to embed graphs. We also use TransR \cite{transr} to
evaluate the effect of a different graph embedding method.
 
In the evaluation, for every sample axiom $C \sqsubseteq D$ or $C
\sqsubseteq \exists R .D$ in the testing set,
we generate predictions for every other axiom $C \sqsubseteq D'$ or $C
\sqsubseteq \exists R .D'$ for every named class $D'$. Then, we
compute the rank of the positive axiom based on the score obtained
from the projected graph. We report mean rank, hits at $\{1, 10, 100\}$
and ROC AUC. We report filtered metrics by not considering the
predictions that exist in the deductive closure. We performed hyperparameter optimization (see
Appendix~\ref{app:hpo}) and provide complete source code for our
experiments at
\url{https://github.com/bio-ontology-research-group/ontology-graph-projections}.

\subsection{Evaluation results}
\label{sec:results}

We test the performance of ontology embeddings in two tasks. First,
we test on prediction of plausible axioms that cannot be inferred but
which may hold true given the other axioms. We use embeddings from
ontologies with removed axioms GO$_{sub}$ and GO$_{ex}$ and evaluate
on axioms $C\sqsubset D$ and $C\sqsubseteq \exists R.D$ existing in the
deductive closure of GO, but not in the deductive
closure of GO$_{sub}$ and GO$_{ex}$, respectively; this task is a form
of ``ontology completion'' \cite{owl2vecstar, context_ont_sub};
we focus on axioms in the deductive closure instead of asserted axioms
to evaluate whether the regularities hold semantically instead of
merely syntactically. 
Second, we test on a deductive
inference task (i.e., test the prediction of axioms in the deductive
closure), similarly to an automated reasoner. We use embeddings from
GO and to compare directly with the first task, we evaluated on the
same test set of axioms used in the first task. Table~\ref{tab:go1} shows the results.

OWL2Vec* can parse complex axioms (i.e.,
$C \equiv D \sqcap \exists R .E$) and generates edges (C,
\text{subclassof}, D) that Onto2Graph does not generate (given the
relational patterns we employ). Furthermore, OWL2Vec* generates
several inverse edges not created by Onto2Graph.
These differences allow OWL2Vec* to rank axioms of type $C \sqsubseteq
D$ higher in Hits@k metrics (See Appendix~\ref{app:projection_analysis}).
The RDF projection generates edges of the form (C, subclassof, D)
where and $C$ or $D$ are not necessarily named classes
but can be blank nodes. The presence on blank nodes adds noise
when embedding RDF graphs, which causes lower values in Hits@k
compared to other methods. 

\begin{table}[t]
  \caption{Results of prediction of axioms of the form
    $C\sqsubseteq D$ and $C \sqsubseteq \exists R. D$ of different
    projection methods on different versions of the Gene Ontology.}
  \label{tab:go1}

  \centering
  \adjustbox{width=\textwidth}{

    \begin{tabular}{lrrrrrrrrrr}
      \toprule
      \multirow{4}{*}{Method} & \multicolumn{10}{c}{Predictions of axioms $C\sqsubseteq D$} \\
      \cmidrule{2-11}
                              & \multicolumn{5}{c|}{$GO$} & \multicolumn{5}{c}{$GO_{sub}$}  \\ \midrule
                              & \multicolumn{1}{c}{MR} & \multicolumn{1}{c}{H@1} & \multicolumn{1}{c}{H@10} & \multicolumn{1}{c}{H@100} & \multicolumn{1}{c}{AUC} & \multicolumn{1}{c}{MR} & \multicolumn{1}{c}{H@1} & \multicolumn{1}{c}{H@10} & \multicolumn{1}{c}{H@100} & \multicolumn{1}{c}{AUC}  \\ \cmidrule{2-11}
                              %& \cntrd{R} & \Cntrd{F} & \cntrd{R} & \Cntrd{F} & \cntrd{R} & \Cntrd{F} & \cntrd{R} & \Cntrd{F} & \cntrd{R} & \Cntrd{F} & \cntrd{R} & \Cntrd{F} & \cntrd{R} & \Cntrd{F} & \cntrd{R} & \Cntrd{F} & \cntrd{R} & \Cntrd{F} & \cntrd{R} & \cntrd{F} \\ \midrule
      Onto2Graph-TransE &  \fst{2947.81} &  0.41   &  0.85    &  12.32   &  \fst{94.23} &  4439.27   &  0.41   &  1.35   &  10.09   &  91.31 \\
      Onto2Graph-TransR &  3599.95   &  0.03   &  0.74    &  6.86    &  92.95   &  \fst{3704.74}   &  0.09   &  0.45   &  8.17    &  \fst{92.75} \\
      OWL2Vec*-TransE   &  4514.51   &  {1.98} &  \fst{10.03} &  \fst{33.87} &  91.16   &  5374.64   &  \fst{1.79} &  \fst{9.55} &  \fst{34.34} &  89.48\\
      OWL2Vec*-TransR   &  3083.20   &  \fst{2.98}   &  5.40    &  22.71   &  93.96   &  4604.52   &  0.01   &  5.90   &  15.24   &  90.98\\
      RDF-TransE        &  4158.77   &  0.36   &  2.85    &  10.03   &  91.86   &  {4122.82} &  0.41   &  2.82   &  10.10   &  {91.93} \\
      RDF-TransR        &  4116.13   &  0.04   &  0.62    &  5.41    &  91.94   &  4899.87   &  0.02   &  0.52   &  6.16    &  90.41 \\
      \midrule
                 & \multicolumn{10}{c}{Predictions of axioms $C \sqsubseteq \exists R.D$} \\
      \cmidrule{2-11}
                 & \multicolumn{5}{c|}{$GO$} & \multicolumn{5}{c}{$GO_{ex}$}  \\ \cmidrule{2-11}
      Onto2Graph-TransE &  9250.77   & \fst{5.21} &  14.26   &  25.53   &  81.93   &  10214.69  &  \fst{4.26} & 13.19   &  22.45   &  80.05\\
      Onto2Graph-TransR &  9430.42   & 0.00   &  6.28    &  7.45    &  81.61   &  10644.37  &  0.00   & 0.00    &  0.43    &  79.23 \\
      OWL2Vec*-TransE   &  \fst{8956.46} & 0.74   &  \fst{21.60} &  \fst{28.09} &  \fst{82.51} &  \fst{9037.64} &  0.21   & \fst{21.17} &  \fst{28.30} &  \fst{82.36} \\
      OWL2Vec*-TransR   &  12746.62  & 0.00   &  0.64    &  4.04    &  75.09   &  13342.04  &  0.00   & 0.53    &  3.09    &  73.93 \\ 
      RDF-TransE        &  12240.94  & 1.49   &  3.51    &  5.74    &  76.08   &  12864.34  &  0.00   & 0.21    &  1.49    &  74.86 \\
      RDF-TransR        &  11976.59  & 0.00   &  0.21    &  4.04    &  76.65   &  10740.08  &  0.00   & 0.00    &  0.00    &  79.07\\
      \bottomrule 
    \end{tabular}
  }
    
\end{table}

For axioms of type $C\sqsubseteq \exists R .D$, Onto2Graph and
OWL2Vec* behave differently. For GO, both methods generate
approximately the same number of
edges (30,266 and 31,429) containing $R$ roles as labels. However,
only around half (18,339) of the edges are shared by both graphs. The
remaining edges for Onto2Graph correspond to those generated by the
reasoning process and for OWL2Vec* correspond to edges generated from
subroles and inverse roles which are ignored by Onto2Graph. Inverse
edges create cycles
in the graph. These differences suffice so that
OWL2Vec* outperforms Onto2Graph in almost all the metrics.

TransR embeds nodes and each relation in different spaces. In OWL2Vec*
projections and Onto2Graph, axioms $C \sqsubseteq D$ and $C\sqsubseteq
\exists R.D$ have the same graph structure ((C,subclassof,D) and
(C,R,D), respectively). TransR helps to differentiate the label
``subclassof'' from other labels in the graph,
improving average ranking metrics such as
Mean Rank and AUC. However, for axioms $C \sqsubseteq \exists R.D$,
where a number of edge labels must be considered, TransR underperforms
compared to TransE for OWL2Vec* and Onto2Graph, while improving for RDF projection.

\begin{table}[tb]
  \caption{Prediction of axioms of the form $C \sqsubseteq \exists
    R. D$ by ranking over  (Case A) 
    axioms $\{C \sqsubseteq
    \exists R. D' | D' \in \mathbf{C}\}$ and (Case B) axioms $\{C \sqsubseteq
    \Box R. D' | D' \in \mathbf{C}, \Box \in \{\exists, \forall \}\}$. We performed this test on
    FoodOn.}
  \label{tab:injectivity}
  \centering

  \begin{tabular}{lrrrrr}
    \toprule
    \multirow{3}{*}{Method} & \multicolumn{5}{c}{Case A} \\ \cmidrule{2-6}
           & \multicolumn{1}{c}{MR} & \multicolumn{1}{c}{H@1} & \multicolumn{1}{c}{H@10} & \multicolumn{1}{c}{H@100} & \multicolumn{1}{c}{AUC} \\ \cmidrule{2-6}
      Onto2Graph-TransE &  \fst{7332.68} &  0.06       &  0.23       &  1.67       &  \fst{78.51} \\
    Onto2Graph-TransR   &  8798.84       &  0.04       &  0.56       &  2.85       &  74.21 \\
    OWL2Vec*-TransE     &  8361.63       &  \fst{0.16} &  \fst{1.46} &  \fst{8.43} &  75.49 \\
    OWL2Vec*-TransR     &  8003.53       &  0.00        &  0.14       &  5.37       &  76.54\\
    RDF-TransE          &  8417.59       &  0.02       &  0.04       &  1.34       &  75.33 \\
    RDF TransR          &  13282.75      &  0.00       &  0.01       &  0.61       &  61.07 \\ \midrule
    & \multicolumn{5}{c}{Case B} \\ \midrule
    Onto2Graph-TransE &  14676.41       &  0.02       &  0.15       &  0.91       &  78.49 \\
    Onto2Graph-TransR &  17609.04       & 0.04        & 0.40         & 2.56        &  74.19   \\
    OWL2Vec*-TransE   &  16732.26       &  \fst{0.08} &  \fst{0.73} &  \fst{5.25} &  75.48 \\
    OWL2Vec*-TransR   &    16017.71     & 0.00        & 0.01        & 2.10        & 76.52 \\
    RDF               &  16605.67       &  0.01       &  0.03       &  0.74       &  75.66 \\
    RDF TransR        &  \fst{14330.19} &  0.00       &  0.00       &  0.08       &  \fst{79.00} \\ \bottomrule
  \end{tabular}

  \end{table}

Additionally, we also evaluated over FoodOn which
contains more complex axioms to determine the effect
of injectivity on predicting axioms, specifically axioms of the type
$C \sqsubseteq \exists R.D $ and $C \sqsubseteq \forall R.D$.  As in
the evaluation of GO, we generated embeddings for FoodOn and evaluated
the performance on the prediction of axioms
$C \sqsubseteq \exists R.D$ existing in the deductive closure of
FoodOn but not in the deductive closure of FoodOn with some axioms
removed (FoodOn$_{sub\_ex}$).

We evaluate the ranking of testing samples $C \sqsubseteq \exists R.D$
among a set of predictions. We have two cases: we rank axioms
$C \sqsubseteq \exists R.D$ among all axioms
$C \sqsubseteq \exists R.D'$ for all named classes $D'$ (case A in
Table~\ref{tab:injectivity}), and, secondly, we rank axioms
$C \sqsubseteq \exists R.D$ among axioms $C \sqsubseteq \Box R.D'$ for
all named class $D'$ and $\Box = \{ \exists, \forall \}$ (case B in
Table~\ref{tab:injectivity}). Non-injective methods (such as OWL2Vec*)
generate multiple axioms when inverting the projection of
$C \sqsubseteq \exists R. D$, making it necessary to consider the
scores of multiple axioms when evaluating axiom inference. We limit
the choice of quantifier to only $\exists$ and $\forall$ and ignore
cardinality restrictions
We also evaluate Onto2Graph projections that are non-injective and
project both $ C\sqsubseteq \exists R.D$ and
$C \sqsubseteq \forall R.D$ onto the same edge (similarly to
OWL2Vec*).

Obviously, methods such as OWL2Vec* and (non-injective) Onto2Graph
decrease the performance from case A to case B. More specifically, the mean rank
doubles because for every axiom $C \sqsubseteq \exists R.D$, another
axiom ($C \sqsubseteq \forall R.D$) has the same score and is ranked
at the same position. In the RDF projection, both
$C \sqsubseteq \exists R.D$ and $C \sqsubseteq \forall R.D$ will,
usually, obtain different scores. Nevertheless, we observe that the
mean ranks almost doubles between case A and B when using TransE. This
is because the subgraphs projected from $C \sqsubseteq \exists R.D$
and $C \sqsubseteq \forall R.D$ differ only in the edge labels
``somevaluesfrom'' and ``allvaluesfrom'' (Appendix
Figure~\ref{fig:rdf_inj}), and, in our case, TransE generates
similar embeddings for ``somevaluesfrom'' and ``allvaluesfrom''. We
also included another experiment using TransR \cite{transr} instead of
TransE; TransR uses a different embedding for relations which are
embedded in a different space than nodes. Although the overall
performance drops in case A, TransR represents relations corresponding
to ``somevaluesfrom'' and ``allvaluesfrom'' differently, illustrated
by an increase in mean rank from case A to case B, and an increase in
AUC.  In the future, further knowledge graph embedding approaches need
to be evaluated. Furthermore, Onto2Graph produces a lower mean rank than
OWL2Vec*. A potential reason is that FoodOn contains complex
axioms that, alike OWL2Vec*, 
Onto2Graph can represent through its reasoning step (see Appendix~\ref{app:projection_analysis}).

\section{Conclusion}

Graph representations of ontologies enable the use of
graph-based machine learning methods to predict axioms. 
Machine learning methods on graphs have been extensively studied, and we
analyzed the properties of different methods that project ontologies
onto graphs and their effects on axiom inference using machine
learning. We find that the properties of graph projections can have a
significant effect on the inference of axioms using ontology
embeddings. Our analysis can be used to further improve graph-based
ontology embeddings and their applications.

\clearpage

\bibliographystyle{plainnat}
\bibliography{refs}

\begin{thebibliography}{36}
\providecommand{\natexlab}[1]{#1}
\providecommand{\url}[1]{\texttt{#1}}
\expandafter\ifx\csname urlstyle\endcsname\relax
  \providecommand{\doi}[1]{doi: #1}\else
  \providecommand{\doi}{doi: \begingroup \urlstyle{rm}\Url}\fi

\bibitem[Allen and Frisch(1982)]{what_is_a_sn}
James~F. Allen and Alan~M. Frisch.
\newblock What’s in a semantic network?
\newblock In \emph{20th Annual Meeting of the Association for Computational
  Linguistics}, page 19–27, Toronto, Ontario, Canada, Jun 1982. Association
  for Computational Linguistics.
\newblock \doi{10.3115/981251.981256}.

\bibitem[Ashburner et~al.(2000)Ashburner, Ball, Blake, Botstein, Butler,
  Cherry, Davis, Dolinski, Dwight, Eppig, Harris, Hill, Issel-Tarver,
  Kasarskis, Lewis, Matese, Richardson, Ringwald, Rubin, and Sherlock]{go}
Michael Ashburner, Catherine~A. Ball, Judith~A. Blake, David Botstein, Heather
  Butler, J.~Michael Cherry, Allan~P. Davis, Kara Dolinski, Selina~S. Dwight,
  Janan~T. Eppig, Midori~A. Harris, David~P. Hill, Laurie Issel-Tarver, Andrew
  Kasarskis, Suzanna Lewis, John~C. Matese, Joel~E. Richardson, Martin
  Ringwald, Gerald~M. Rubin, and Gavin Sherlock.
\newblock Gene ontology: tool for the unification of biology.
\newblock \emph{Nature Genetics}, 25\penalty0 (1):\penalty0 25--29, May 2000.
\newblock \doi{10.1038/75556}.

\bibitem[{Baader} et~al.(2003){Baader}, {Calvanese}, {McGuinness}, {Nardi}, and
  {Patel-Schneider}]{DLhandbook}
Franz {Baader}, Diego {Calvanese}, Deborah {McGuinness}, Daniele {Nardi}, and
  Peter~F. {Patel-Schneider}, editors.
\newblock \emph{The Description Logic Handbook: Theory, Implementation, and
  Applications}.
\newblock Cambridge University Press, 2003.

\bibitem[Bordes et~al.(2013)Bordes, Usunier, Garcia-Duran, Weston, and
  Yakhnenko]{transe}
Antoine Bordes, Nicolas Usunier, Alberto Garcia-Duran, Jason Weston, and Oksana
  Yakhnenko.
\newblock Translating embeddings for modeling multi-relational data.
\newblock In \emph{Advances in Neural Information Processing Systems},
  volume~26. Curran Associates, Inc., 2013.

\bibitem[Carroll et~al.(2012)Carroll, Herman, and Patel-Schneider]{rdf_owl2}
Jeremy Carroll, Ivan Herman, and Peter Patel-Schneider.
\newblock Owl 2 web ontology language rdf-based semantics (second edition),
  2012.

\bibitem[Chen et~al.(2021{\natexlab{a}})Chen, Hu, Jimenez-Ruiz, Holter,
  Antonyrajah, and Horrocks]{owl2vecstar}
Jiaoyan Chen, Pan Hu, Ernesto Jimenez-Ruiz, Ole~Magnus Holter, Denvar
  Antonyrajah, and Ian Horrocks.
\newblock {OWL2Vec*}: embedding of {OWL} ontologies.
\newblock \emph{Machine Learning}, June 2021{\natexlab{a}}.
\newblock \doi{10.1007/s10994-021-05997-6}.

\bibitem[Chen et~al.(2021{\natexlab{b}})Chen, Jim{\'e}nez-Ruiz, Horrocks,
  Antonyrajah, Hadian, and Lee]{owl2vec_align}
Jiaoyan Chen, Ernesto Jim{\'e}nez-Ruiz, Ian Horrocks, Denvar Antonyrajah, Ali
  Hadian, and Jaehun Lee.
\newblock Augmenting ontology alignment by semantic embedding and distant
  supervision.
\newblock In \emph{Extended Semantic Web Conference}, 2021{\natexlab{b}}.

\bibitem[Chen et~al.(2022)Chen, He, Geng, Jimenez-Ruiz, Dong, and
  Horrocks]{context_ont_sub}
Jiaoyan Chen, Yuan He, Yuxia Geng, Ernesto Jimenez-Ruiz, Hang Dong, and Ian
  Horrocks.
\newblock Contextual semantic embeddings for ontology subsumption prediction,
  2022.

\bibitem[Chen et~al.(2020{\natexlab{a}})Chen, Althagafi, and Hoehndorf]{dl2vec}
Jun Chen, Azza Althagafi, and Robert Hoehndorf.
\newblock Predicting candidate genes from phenotypes, functions and anatomical
  site of expression.
\newblock \emph{Bioinformatics}, 37\penalty0 (6):\penalty0 853--860, October
  2020{\natexlab{a}}.
\newblock \doi{10.1093/bioinformatics/btaa879}.

\bibitem[Chen et~al.(2020{\natexlab{b}})Chen, Wang, Zhao, Cheng, Zhao, and
  Duan]{kg_completion}
Zhe Chen, Yuehan Wang, Bin Zhao, Jing Cheng, Xin Zhao, and Zongtao Duan.
\newblock Knowledge graph completion: A review.
\newblock \emph{IEEE Access}, 8:\penalty0 192435--192456, 2020{\natexlab{b}}.
\newblock \doi{10.1109/ACCESS.2020.3030076}.

\bibitem[Corbett(2008)]{conceptual_ontology}
Dan~R. Corbett.
\newblock \emph{Graph-Based Representation and Reasoning for Ontologies}, page
  351–379.
\newblock Springer Berlin Heidelberg, Berlin, Heidelberg, 2008.
\newblock ISBN 978-3-540-78293-3.
\newblock \doi{10.1007/978-3-540-78293-3_8}.

\bibitem[Dau(2005)]{dau_2005}
F.~Dau.
\newblock Mathematical logic with diagrams based on the existential graphs of
  peirce.
\newblock 2005.

\bibitem[Dau(2001)]{Dau_2001}
Frithjof Dau.
\newblock Concept graphs and predicate logic.
\newblock In Harry~S. Delugach and Gerd Stumme, editors, \emph{Conceptual
  Structures: Broadening the Base}, page 72–86, Berlin, Heidelberg, 2001.
  Springer Berlin Heidelberg.
\newblock ISBN 978-3-540-44583-8.

\bibitem[Dau and Eklund(2007)]{diagrammatic_alc}
Frithjof Dau and Peter Eklund.
\newblock A diagrammatic reasoning system for $\mathcal{ALC}$.
\newblock In Zili Zhang and Jörg Siekmann, editors, \emph{Knowledge Science,
  Engineering and Management}, page 39–51, Berlin, Heidelberg, 2007. Springer
  Berlin Heidelberg.
\newblock ISBN 978-3-540-76719-0.

\bibitem[Dooley et~al.(2018)Dooley, Griffiths, Gosal, Buttigieg, Hoehndorf,
  Lange, Schriml, Brinkman, and Hsiao]{foodon}
Damion~M. Dooley, Emma~J. Griffiths, Gurinder~S. Gosal, Pier~L. Buttigieg,
  Robert Hoehndorf, Matthew~C. Lange, Lynn~M. Schriml, Fiona S.~L. Brinkman,
  and William W.~L. Hsiao.
\newblock {FoodOn}: a harmonized food ontology to increase global food
  traceability, quality control and data integration.
\newblock \emph{npj Science of Food}, 2\penalty0 (1), December 2018.
\newblock \doi{10.1038/s41538-018-0032-6}.

\bibitem[Golbreich and Horrocks(2007)]{owl1_1}
Christine Golbreich and Ian Horrocks.
\newblock The obo to owl mapping, go to owl 1.1!
\newblock In \emph{OWL: Experiences and Directions}, 2007.

\bibitem[Grau et~al.(2008)Grau, Horrocks, Motik, Parsia, Patel-Schneider, and
  Sattler]{owl2}
Bernardo~Cuenca Grau, Ian Horrocks, Boris Motik, Bijan Parsia, Peter
  Patel-Schneider, and Ulrike Sattler.
\newblock Owl 2: The next step for owl.
\newblock \emph{Journal of Web Semantics}, 6\penalty0 (4):\penalty0 309–322,
  Nov 2008.
\newblock ISSN 1570-8268.
\newblock \doi{10.1016/j.websem.2008.05.001}.

\bibitem[Grohe(2007)]{logicGA}
Martin Grohe.
\newblock Logic, graphs, and algorithms.
\newblock \emph{Electron. Colloquium Comput. Complex.}, TR07, 2007.

\bibitem[Hoehndorf et~al.(2010)Hoehndorf, Oellrich, Dumontier, Kelso,
  Rebholz-Schuhmann, and Herre]{relations_as_patterns}
Robert Hoehndorf, Anika Oellrich, Michel Dumontier, Janet Kelso, Dietrich
  Rebholz-Schuhmann, and Heinrich Herre.
\newblock Relations as patterns: bridging the gap between {OBO} and {OWL}.
\newblock \emph{{BMC} Bioinformatics}, 11\penalty0 (1), August 2010.
\newblock \doi{10.1186/1471-2105-11-441}.
\newblock URL \url{https://doi.org/10.1186/1471-2105-11-441}.

\bibitem[Horrocks()]{obo_syntax_semantics}
Ian Horrocks.
\newblock Obo flat file format syntax and semantics and mapping to owl web
  ontology language.
\newblock URL \url{http://www.cs.man.ac.uk/~horrocks/obo/}.

\bibitem[Horrocks et~al.(2006)Horrocks, Kutz, and Sattler]{sroiq}
Ian Horrocks, Oliver Kutz, and Ulrike Sattler.
\newblock The even more irresistible sroiq.
\newblock In \emph{International Conference on Principles of Knowledge
  Representation and Reasoning}, 2006.

\bibitem[Kazakov et~al.(2013)Kazakov, Krötzsch, and Simančík]{elk}
Yevgeny Kazakov, Markus Krötzsch, and František Simančík.
\newblock The incredible elk.
\newblock \emph{Journal of Automated Reasoning}, 53\penalty0 (1):\penalty0
  1–61, Nov 2013.
\newblock ISSN 1573-0670.
\newblock \doi{10.1007/s10817-013-9296-3}.

\bibitem[Kulmanov et~al.(2020)Kulmanov, Smaili, Gao, and
  Hoehndorf]{semantic_similarity}
Maxat Kulmanov, Fatima~Zohra Smaili, Xin Gao, and Robert Hoehndorf.
\newblock {Semantic similarity and machine learning with ontologies}.
\newblock \emph{Briefings in Bioinformatics}, 22\penalty0 (4), 10 2020.
\newblock ISSN 1477-4054.
\newblock \doi{10.1093/bib/bbaa199}.
\newblock bbaa199.

\bibitem[Lin et~al.(2015)Lin, Liu, Sun, Liu, and Zhu]{transr}
Yankai Lin, Zhiyuan Liu, Maosong Sun, Yang Liu, and Xuan Zhu.
\newblock Learning entity and relation embeddings for knowledge graph
  completion.
\newblock In \emph{AAAI Conference on Artificial Intelligence}, 2015.

\bibitem[Ren and Leskovec(2020)]{betaE}
Hongyu Ren and Jure Leskovec.
\newblock Beta embeddings for multi-hop logical reasoning in knowledge graphs.
\newblock \emph{ArXiv}, abs/2010.11465, 2020.

\bibitem[Rodr{\'{\i}}guez-Garc{\'{\i}}a and Hoehndorf(2018)]{onto2graph}
Miguel~{\'{A}}ngel Rodr{\'{\i}}guez-Garc{\'{\i}}a and Robert Hoehndorf.
\newblock Inferring ontology graph structures using {OWL} reasoning.
\newblock \emph{{BMC} Bioinformatics}, 19\penalty0 (1), January 2018.
\newblock \doi{10.1186/s12859-017-1999-8}.

\bibitem[Smith et~al.(2005)Smith, Ceusters, Klagges, K\"{o}hler, Kumar, Lomax,
  Mungall, Neuhaus, Rector, and Rosse]{Smith2005}
Barry Smith, Werner Ceusters, Bert Klagges, Jacob K\"{o}hler, Anand Kumar, Jane
  Lomax, Chris Mungall, Fabian Neuhaus, Alan~L Rector, and Cornelius Rosse.
\newblock \emph{Genome Biology}, 6\penalty0 (5):\penalty0 R46, 2005.
\newblock \doi{10.1186/gb-2005-6-5-r46}.

\bibitem[Smith et~al.(2007)Smith, Ashburner, Rosse, Bard, Bug, Ceusters,
  Goldberg, Eilbeck, Ireland, Mungall, Leontis, Rocca-Serra, Ruttenberg,
  Sansone, Scheuermann, Shah, Whetzel, and Lewis]{obofoundry}
Barry Smith, Michael Ashburner, Cornelius Rosse, Jonathan Bard, William Bug,
  Werner Ceusters, Louis~J Goldberg, Karen Eilbeck, Amelia Ireland,
  Christopher~J Mungall, Neocles Leontis, Philippe Rocca-Serra, Alan
  Ruttenberg, Susanna-Assunta Sansone, Richard~H Scheuermann, Nigam Shah,
  Patricia~L Whetzel, and Suzanna Lewis.
\newblock The obo foundry: coordinated evolution of ontologies to support
  biomedical data integration.
\newblock \emph{Nature biotechnology}, 25\penalty0 (11):\penalty0 1251, Nov
  2007.
\newblock ISSN 1087-0156.
\newblock \doi{10.1038/nbt1346}.

\bibitem[Sowa(2011)]{pierce_tutorial}
John~F. Sowa.
\newblock Peirce's tutorial on existential graphs.
\newblock \emph{Semiotica}, 2011\penalty0 (186), January 2011.
\newblock \doi{10.1515/semi.2011.060}.

\bibitem[Sowa(2013)]{from_eg_to_cg}
John~F. Sowa.
\newblock From existential graphs to conceptual graphs.
\newblock \emph{International Journal of Conceptual Structures and Smart
  Applications}, 1\penalty0 (1):\penalty0 39--72, January 2013.
\newblock \doi{10.4018/ijcssa.2013010103}.

\bibitem[Tang et~al.(2022{\natexlab{a}})Tang, Hinnerichs, Peng, Zhang, and
  Hoehndorf]{falcon}
Zhenwei Tang, Tilman Hinnerichs, Xi~Peng, Xiangliang Zhang, and Robert
  Hoehndorf.
\newblock Falcon: Faithful neural semantic entailment over alc ontologies,
  2022{\natexlab{a}}.

\bibitem[Tang et~al.(2022{\natexlab{b}})Tang, Pei, Peng, Zhuang, Zhang, and
  Hoehndorf]{tar}
Zhenwei Tang, Shichao Pei, Xi~Peng, Fuzhen Zhuang, Xiangliang Zhang, and Robert
  Hoehndorf.
\newblock Tar: Neural logical reasoning across tbox and abox,
  2022{\natexlab{b}}.

\bibitem[Wang et~al.(2017)Wang, Mao, Wang, and Guo]{kgsurvey}
Quan Wang, Zhendong Mao, Bin Wang, and Li~Guo.
\newblock Knowledge graph embedding: A survey of approaches and applications.
\newblock \emph{{IEEE} Transactions on Knowledge and Data Engineering},
  29\penalty0 (12):\penalty0 2724--2743, December 2017.
\newblock \doi{10.1109/tkde.2017.2754499}.

\bibitem[Woods(1975)]{what_is_a_link}
William~A. Woods.
\newblock \emph{WHAT’S IN A LINK: Foundations for Semantic Networks}, page
  35–82.
\newblock Morgan Kaufmann, San Diego, Jan 1975.
\newblock ISBN 978-0-12-108550-6.
\newblock \doi{10.1016/B978-0-12-108550-6.50007-0}.

\bibitem[Yang et~al.(2022)Yang, Qing, Li, Lu, and Lin]{gammaE}
D.~Yang, Peijun Qing, Y.~Li, H.~Lu, and Xiaodong Lin.
\newblock Gammae: Gamma embeddings for logical queries on knowledge graphs.
\newblock In \emph{Conference on Empirical Methods in Natural Language
  Processing}, 2022.

\bibitem[Zhapa-Camacho et~al.(2022)Zhapa-Camacho, Kulmanov, and
  Hoehndorf]{mowl}
Fernando Zhapa-Camacho, Maxat Kulmanov, and Robert Hoehndorf.
\newblock {mOWL: Python library for machine learning with biomedical
  ontologies}.
\newblock \emph{Bioinformatics}, 12 2022.
\newblock ISSN 1367-4803.
\newblock \doi{10.1093/bioinformatics/btac811}.
\newblock btac811.

\end{thebibliography}

\appendix
\newpage
\section{Analysis of the performance of projection methods}
\label{app:projection_analysis}

As shown in Section~\ref{sec:results}, in most cases OWL2Vec* obtains
higher values on Hits@1 but Onto2Graph obtains lower mean rank. This
phenomenon is due to the different capabilities of each method.
We show the following example in GO using the class $\mbox{GO}\_2000859$,
which is involved in the following axiom in the training set:

\begin{equation}
  \label{eq:axiom_in_go}
  \mbox{GO}\_2000859 \equiv \mbox{GO}\_0065007 \sqcap \exists
  \mbox{RO}\_0002212. \mbox{GO}\_0035932
\end{equation}

And is involved in the following axiom in the testing set:

\begin{equation}
  \label{eq:testing_axiom_in_go}
  \mbox{GO}\_2000859 \sqsubseteq \mbox{GO}\_0023051
\end{equation}

OWL2Vec*  generates an edge $(\mbox{GO}\_2000859,  \mbox{subclassof},
\mbox{GO}\_0065007)$ from axiom~\ref{eq:axiom_in_go} but Onto2Graph
does not.
Furthermore, classes $\mbox{GO}\_0065007$ and $\mbox{GO}\_0023051$ 
are involved in training axioms, and the edge generated by OWL2Vec* is
important in the prediction of the testing
axiom~\ref{eq:testing_axiom_in_go} and contributing to the high value
at Hits@k.
  
Furthermore, RDF performs worse than other methods because of the
noise introduced by blank nodes. For example,
axiom~\ref{eq:axiom_in_go}, will be projected as: $(\mbox{GO}\_2000859, \mbox{subclassof}, \mbox{m})$,
$(\mbox{m}, \mbox{intersection}, \mbox{l})$, $(\mbox{l},
\mbox{first}, \mbox{GO}\_0065007)$.

In the case of FoodOn, we notice that Onto2Graph obtains much lower
mean rank than OWL2Vec*. This happens because Onto2Graph, through its
reasoning step, generates edges than OWL2Vec* cannot generate due to
the complexity of the axioms in FoodOn. For example, from the
following axiom involving the entity $\mbox{CDNO}\_0200195$:

\[
\mbox{CDNO}\_0200195 \equiv \mbox{PATO}\_0000033 \sqcap \exists \mbox{RO}\_0000052. (\mbox{CHEBI}\_12777 \sqcap \exists \mbox{BFO}\_0000050 . \mbox{BFO}\_0000040) 
\]
  
Onto2Graph generates
\[ (\mbox{CDNO}\_0200195, \mbox{RO}\_0000052,
  \mbox{CHEBI}\_12777) \]
whereas OWL2Vec* does not generate any edge with roles as edge labels. This difference between both graphs enables Onto2Graph to get lower mean rank
when predicting axioms $C \sqsubseteq \exists R.D$.

\section{Description logics and ontologies}
\label{app:ontologies}
Ontologies can be constructed using Description Logics.  A Description
Logic (DL) \cite{DLhandbook} theory is defined over a signature
$\Sigma = (\mathbf{C}, \mathbf{R}, \mathbf{I})$ where $\mathbf{C}$ is
a set of class names, $\mathbf{R}$ a set of role names, $\mathbf{I}$
a set of individual names. There are several description logic
languages that differ from each other on the operators that they
support. In the description logic $\mathcal{ALC}$, a concept
description is constructed inductively from class names using the
operations of negation ($\neg$), intersection ($\sqcap$), union
($\sqcup$), existential ($\exists$) and universal quantification
($\forall$).

In DLs, subsumption axioms between concept descriptions can be defined
using the subsumption operation ($\sqsubseteq$). To define the semantics of a DL, we need
an interpretation domain $\Delta^{\mathcal{I}}$ and an interpretation
function $ \cdot^{\mathcal{I}}$. In $\mathcal{ALC}$, for a class
name $A \in \mathbf{C}$, its interpretation is the set
$A^{\mathcal{I}} \subseteq \Delta^{\mathcal{I}}$. The semantics of
concept descriptions is constructed inductively:

\begin{equation}
\begin{aligned}
  {\perp}^{\mathcal{I}} &=\emptyset \\
  {\top}^{\mathcal{I}} &=\Delta^{\mathcal{I}} \\
  (\neg A)^{\mathcal{I}} &=\Delta^{\mathcal{I}} \backslash A^{\mathcal{I}} \\
  (C \sqcap D)^{\mathcal{I}} &=C^{\mathcal{I}} \cap D^{\mathcal{I}} \\
  (\forall R . C)^{\mathcal{I}} &=\left\{a \in \Delta^{\mathcal{I}}
    \mid \forall b .(a, b) \in R^{\mathcal{I}} \rightarrow b \in
    C^{\mathcal{I}}\right\} \\
  (\exists R . \top)^{\mathcal{I}} &=\left\{a \in \Delta^{\mathcal{I}}
    \mid \exists b .(a, b) \in R^{\mathcal{I}}\right\}.
\end{aligned}
\end{equation}

\section{Details on  implementations of projection methods}
\label{app:owl2vec}

For Onto2Graph, we relied on the original implementation of \cite{onto2graph} found at \url{https://github.com/bio-ontology-research-group/Onto2Graph}. In the case of RDF, we used the Python library \texttt{rdflib}, found at \url{https://github.com/RDFLib/rdflib}

In the case of the projection found in OWL2Vec*, we used the implementation found in
mOWL\cite{mowl}. Both mOWL and the original implementation of OWL2Vec*
projection, project axioms with complex superclasses such as
$C \sqsubseteq D \sqcap E$. We added this rule in Table~\ref{tab:owl2vec}.

\begin{table}[ht]
  \caption{Projection rules for OWL2Vec* model.}
  \label{tab:owl2vec}
  \begin{tabular}{|p{2.5cm}|l|l|}
\hline
  Axiom or triple(s) of condition 1          & Axiom or triple(s) of condition 2                                                          & Projected triple(s)                                                    \\ \hline
  $A       \sqsubseteq \Box r.D$             & \multirow{3}{*}{$D \equiv B |B_1 \sqcup \ldots \sqcup B_n|  B_1 \sqcap \ldots \sqcap B_n$} & \multirow{3}{*}{$\langle A, r, B\rangle$ for $i \in 1, \ldots, n$}                              \\ \cline{1-1}
  or                                         &                                                                                            &                                                                        \\ \cline{1-1}
  $\Box r.D \sqsubseteq A$                   &                                                                                            &                                                                        \\ \hline
  $\exists r. \top \sqsubseteq A (domain)$   & $\top \sqsubseteq \forall     r . B$ (range)                                               & \multirow{5}{*}{$\langle A, r, B_{i}\rangle$} \\ \cline{1-2}
  $A \sqsubseteq \exists r . {b}$            & $B(b)$                                                                                     &                                                                        \\ \cline{1-2}
  $r \sqsubseteq r'$                         & $\langle A, r', B \rangle$ has been projected                                              &                                                                        \\ \cline{1-2}
         $r' \sqsubseteq r^{-1}$             & $\langle B, r', A \rangle$ has been projected                                              &                                                                        \\\cline{1-2}
  $s_1 \circ \ldots \circ s_n \sqsubseteq r$ & $\langle A, s_1, C_1 \rangle$ \ldots $\langle C_n, s_n, C_B \rangle$ has been projected    &                                                                        \\ \hline
    \multirow{2}{*}{$B \sqsubseteq A$}         & \multirow{2}{*}{--}                                                                        & $\langle B, \mbox{ subClassOf}, A\rangle$                                \\ \cline{3-3}
                                             &                                                                                            & $\langle A,  \mbox{subClassOf}^{-1}, B\rangle$                           \\ \hline
    \multirow{2}{*}{$A(a)$}                    & \multirow{2}{*}{--}                                                                        & $\langle a, \mbox{ type}, A\rangle$                                      \\ \cline{3-3}
                                             &                                                                                            & $\langle A, \mbox{ type}^{-1}, a\rangle$                                 \\ \hline
  $r(a,b)$                                   & --
                                                                                                                                          & $\langle a,r,b \rangle$                                                \\ \hline
    \multicolumn{3}{l}{New rule} \\
    \hline
    $A \sqsubseteq B \sqcap \exists R. C$ & & $\langle A, \mbox{subClassOf}, B \rangle$ \\ \hline
\end{tabular}

\end{table}

\newpage
\section{Examples of graphs generated by different projection methods}
\label{app:graphs}

\begin{figure}[ht]
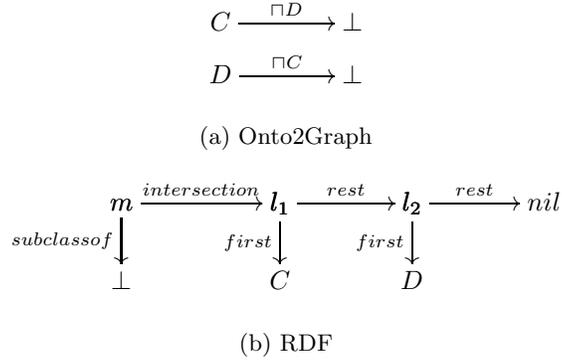

\centering
\begin{subfigure}{.2\linewidth}
  \centering
  \begin{equation*}
    \bfig
    \morphism[C`\bot;\sqcap D]
    \morphism(0,-200)[D`\bot;\sqcap C]
    \efig
  \end{equation*}
  \caption{Onto2Graph}
  \label{fig:onto1}
\end{subfigure}
    \hfill
\begin{subfigure}{.8\linewidth}
  \centering
    \begin{equation*}
    \bfig
    \morphism(-100,0)<600,0>[m`l_1;intersection]
    \morphism(500,0)<500,0>[l_1`l_2;rest]
    \morphism(1000,0)<500,0>[l_2`nil;rest]
    \morphism(-100,0)<0,-300>[m`\bot;subclassof]
    \morphism(500,0)<0,-300>[l_1`C;first]
    \morphism(1000,0)<0,-300>[l_2`D;first]
    \efig
  \end{equation*}
  \caption{RDF}\label{fig:rdf1}
\end{subfigure}
 \caption{Graphs generated by each projection method on the axiom $C
   \sqcap D \sqsubseteq \bot$. This axiom is applicable for
   Onto2Graph (if the appropriate pattern is defined) and for RDF projection.}\label{fig:axiom1}
\end{figure}

\begin{figure}[ht]
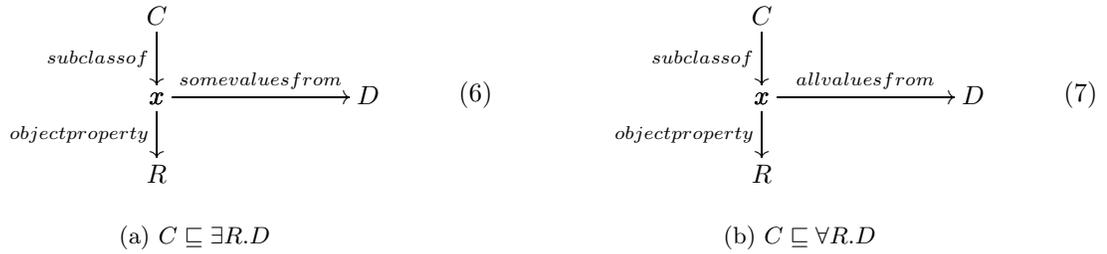

\centering
\begin{subfigure}{0.48\linewidth}
  \centering
   \begin{equation}
    \bfig
    \morphism(-1000,0)<800,0>[x`D;somevaluesfrom]
    \morphism(-1000,300)<0,-300>[C`x;subclassof]
    \morphism(-1000,0)<0,-300>[x`R;objectproperty]
    \efig
  \end{equation}
  \caption{$C \sqsubseteq \exists R. D$}
\end{subfigure}
\begin{subfigure}{0.48\linewidth}
  \centering
   \begin{equation}
    \bfig
    \morphism(-1000,0)<800,0>[x`D;allvaluesfrom]
    \morphism(-1000,300)<0,-300>[C`x;subclassof]
    \morphism(-1000,0)<0,-300>[x`R;objectproperty]
    \efig
  \end{equation}
  \caption{$C \sqsubseteq \forall R. D$}
\end{subfigure} 
\caption{RDF projection of axioms $C \sqsubseteq \exists R. D$ and $C \sqsubseteq \forall R. D$}\label{fig:rdf_inj}
\end{figure}

To show the differences between the different projection results,
Figure~\ref{fig:axiom1} shows the subgraphs generated by each method
on the axiom $C \sqcap D \sqsubseteq \bot$.
Similarly, Figure~\ref{fig:rdf_inj} shows the projections of axioms $C
\sqsubseteq \exists R .D$ and $C \sqsubseteq \forall R .D$ using the
RDF projection.

% \begin{figure}[ht]
% \centering
% \begin{subfigure}{.3\linewidth}
%   \centering
%   \begin{equation*}
%     \bfig
%     \morphism[C`D;R]
%     \morphism(0,-200)[C`E;R]
%     \efig
%   \end{equation*}
%   \caption{DL2Vec}
%   \label{fig:d2v2}
% \end{subfigure}
%     \hfill
% \begin{subfigure}{.3\linewidth}
%   \centering
%     \begin{equation*}
%     \bfig
%     \morphism[C`D;R]
%     \morphism(0,-200)[C`E;R]
%     \efig
%   \end{equation*}
%   \caption{OWL2Vec*}
%   \label{fig:o2v2}
% \end{subfigure}
%    \hfill
% \begin{subfigure}{.3\linewidth}
%   \centering
%   \begin{equation}
%     \bfig
%     \morphism[C`X_i;R]
%     \efig
%   \end{equation}
%   \caption{Onto2Graph}
%   \label{fig:onto2}
% \end{subfigure}
% \bigskip
% \begin{subfigure}{\linewidth}
%   \centering
%    \begin{equation}
%     \bfig
%     \morphism(-1000,0)<800,0>[x`m;somevaluesfrom]
%     \morphism(-200,0)<700,0>[m`l_1;intersection]
%     \morphism(500,0)<500,0>[l_1`l_2;rest]
%     \morphism(1000,0)<500,0>[l_2`nil;rest]
%     \morphism(-1000,300)<0,-300>[C`x;subclassof]
%     \morphism(-1000,0)<0,-300>[x`R;objectproperty]
%     \morphism(500,0)<0,-300>[l_1`D;first]
%     \morphism(1000,0)<0,-300>[l_2`E;first]
%     \efig
%   \end{equation}
%   \caption{RDF}
%   \label{fig:image3}
% \end{subfigure} 
% \caption{Graphs generated by each projection method on the axiom $C
%   \sqsubseteq \exists R. (D\sqcap E)$. For the case of Onto2Graph, we
%   can query every class $X_i\equiv D \sqcap E$ and generate the
%   corresponding edge.}\label{fig:axiom2}
% \end{figure}

\newpage
\section{Graphs generated for GO and FOODON}
\label{app:use_cases}

For the quantitative analysis of projection methods, we chose two
ontologies: GO and FoodOn. Table~\ref{tab:number_edges} shows the number
of edges generated by each projection method on the different
ontologies.

\begin{table}[ht]
  \centering
  \caption{Number of edges generated by projection methods. $\sqsubseteq$ represents the edges for subclass axioms between named classes (Onto2Graph, OWL2Vec*, RDF) or between a named class and a blank node (RDF). $\sqsubseteq^{-1}$ represents the inverse edge of $\sqsubseteq$.}\label{tab:number_edges}
  \adjustbox{width=\textwidth}{
    \begin{tabular}{crrr|rrr|rrr|rrr}
      \toprule
      \multirow{2}{*}{Projection method} & \multicolumn{12}{c}{Edges generated on each ontology} \\ \cmidrule{2-13}
                                         & \multicolumn{3}{c|}{$GO$}   & \multicolumn{3}{c|}{$GO_{sub}$} & \multicolumn{3}{c|}{$GO_{ex}$} & \multicolumn{3}{c}{$FoodOn$}  \\ \midrule
                                         & Total & $\sqsubseteq$ & $\sqsubseteq^{-1}$ & Total & $\sqsubseteq$ & $\sqsubseteq^{-1}$ & Total & $\sqsubseteq$ & $\sqsubseteq^{-1}$ & Total & $\sqsubseteq$ & $\sqsubseteq^{-1}$ \\ \cmidrule{2-13} 
      Onto2Graph &  99619 &  69353(70\%) &    -- & 97660 & 66886(68\%)  &    -- &  96294 & 69353(72\%) &   --  &  77203 & 36755(47\%)  & --\\ 
      OWL2Vec*   & 189849 &  79210(42\%) & 79210(42\%) & 176281 & 72426(41\%) & 72426(41\%) & 188355 & 79210(42\%) & 79210(42\%) &  89980 & 40248(44\%) & 40248(44\%) \\ 
      RDF        & 709409 &  87602(12\%) &    -- & 702477 & 80667(11\%) &   --  & 702128 & 85781(12\%) &   --  & 210718 & 46668(22\%) & --\\ \bottomrule
    \end{tabular}

  }
  
\end{table}

\begin{table}[ht]
  \centering
  \caption{Number of edges projected from the FoodOn ontology from axioms $C \sqsubseteq \exists R. D$ for methods Onto2Graph and OWL2Vec*. ``With self-loops'' represent edges of the from (C, R, C) for some ontology role R. ``Shared'' represent the number of edges found in both graphs.}\label{tab:foodon_edges}
  \begin{tabular}{lrrrr}
    \toprule
    \multirow{2}{*}{Projection method} & \multicolumn{4}{c}{Number of existential axioms $C \sqsubseteq \exists R .D$} \\ \cmidrule{2-5}
                                       & Total & With self-loops & Shared & Not-shared \\ \midrule
    Onto2Graph & 40448 & 32018 & 6583 & 1847 \\
    OWL2Vec*   &  9484 & 4 & 6583 & 2899 \\\bottomrule
  \end{tabular}
\end{table}

\section{Knowledge graph embedding method TransE}
\label{app:kge}

In TransE\cite{transe}, every edge $(h,r,t)$ is assigned a distance score
\begin{equation}
  \label{eq:transe}
  d_{(h,r,t)}= ||h + r -t||
\end{equation}

That is, every edge label is considered as a translation betweeon the head and tail nodes.
The training objective is denoted as $\mathcal{L}$:

\begin{equation}
  \label{eq:transe_loss}
  \mathcal{L} = \max(0, d_{(h,r,t)} - d_{(h,r,t')} + \gamma)
\end{equation}

where $(h,r,t)$ is a \emph{positive triple} that exists in the graph
and $(h,r,t')$ is a \emph{negative triple} that does not exist in the
graph and is computed by corrupting the node $t$ by another node that
is chosen randomly. $\mathcal{L}$ tries to minimize distance of
positive triples with respect to negative ones. $\gamma$ is a margin
parameter that enforces a minimum separation between the scores of a
positive and a negative sample.

\section{Hyperparameter optimization}
\label{app:hpo}
For all the methods, we performed hyperparameter optimization of the
following parameters: embedding size $[64, 128, 256]$, margin
($\gamma$) $[0.0, 0.2, 0.4]$, L2 regularization factor
$[0.0, 1e^{-4}, 5e^{-4}]$, batch size $[4096, 8192, 16384]$ and
learning rate $[0.1, 0.01, 0.001]$.

\begin{table}[ht]
  \caption{Hyperparameters chosen for projection methods in
    experiments regarding Table~\ref{tab:go1}}\label{tab:hpo_go1}
  \centering
  \begin{tabular}{lrrrrr}
    \toprule
    \multirow{2}{*}{Method} & \multicolumn{5}{c}{Chosen hyperparameter values}            \\ \cmidrule{2-6}
                            & Dimension & Margin & L2 Reg. & Batch size & Learning rate   \\\midrule
                            & \multicolumn{5}{c}{Prediction of $C \sqsubseteq D$ with GO and GO$_{sub}$} \\\midrule
    Onto2Graph-TransE & 128 & 0.4 & 0.0005 & 4096 & 0.010 \\
    Onto2Graph-TransR & 128 & 0.4 & 0.0001 & 8192 & 0.001  \\
    OWL2Vec*-TransE   & 128 & 0.2 & 0.0001 & 8192 & 0.010 \\
    OWL2Vec*-TransR   & 128 & 0.4 & 0.0001 & 8192 & 0.001 \\
    RDF-TransE        & 64  & 0.2 & 0.0    & 4096 & 0.010 \\
    RDF-TransR        & 128 & 0.4 & 0.0    & 8192 & 0.001 \\ \midrule

                      & \multicolumn{5}{c}{Prediction of $C \sqsubseteq \exists R. D$ with GO and GO$_{ex}$}         \\ \midrule
    Onto2Graph-TransE &  64  & 0.4 & 0.0    & 8192  & 0.100 \\
    Onto2Graph-TransR & 256 & 0.4 & 0.0001 & 4096 & 0.001 \\
    OWL2Vec*-TransE   & 128  & 0.4 & 0.0    & 16384 & 0.001 \\
    OWL2Vec*-TransR   &  64 & 0.0 & 0.0 & 4096 & 0.001 \\
    RDF-TransE        &  64  & 0.4 & 0.0    & 8192  & 0.010 \\
    RDF-TransR        &  64 & 0.4 & 0.0005 & 16384 & 0.001 \\
    \bottomrule
    
  \end{tabular}
\end{table}

\begin{table}[ht]
  \caption{Hyperparameters chosen for projection methods in
    experiments regarding Table~\ref{tab:injectivity}}\label{tab:hpo_injectivity}
  \centering
  \begin{tabular}{crrrrr}
    \toprule
    \multirow{2}{*}{Method} & \multicolumn{5}{c}{Chosen hyperparameter values}            \\ \cmidrule{2-6}
                            & Dimension & Margin & L2 Reg. & Batch size & Learning rate   \\\midrule
                            & \multicolumn{5}{c}{Case A and Case B} \\ \midrule
    Onto2Graph-TransE & 64.0 & 0.0 & 0.0 & 16384.0 & 0.010 \\
    Onto2Graph-TransR & 256 & 0.2 & 0.0005 & 16384 & 0.001 \\
    OWL2Vec*-TransE   & 256.0 & 0.4 & 0.0001 & 16384.0 & 0.001 \\
    OWL2Vec*-TransR & 64 & 0.2 & 0.0001 & 16384 &0.001 \\
    RDF-TransE        & 128.0 & 0.0 & 0.0 & 4096.0 & 0.001 \\ 
    RDF-TransR        & 128.0 & 0.0 & 0.0 & 4096.0 & 0.001 \\\bottomrule
  \end{tabular}
\end{table}

\end{document}